\pdfoutput=1
\documentclass[sigconf,nonacm]{acmart}

\usepackage{graphicx}
\usepackage{booktabs}
\usepackage{amsmath}
\usepackage{microtype}
\usepackage{url}

\renewcommand\footnotetextcopyrightpermission[1]{}

\title{BanglaDial-Abuse: A Corpus-Grounded Dataset for Regional Dialect Identification in Abusive Bangla Text}

\author{Hasin Almas Sifat}
\affiliation{%
  \institution{Department of Computer Science}
  \institution{American International University-Bangladesh}
  \city{Dhaka}
  \country{Bangladesh}
}
\email{22-48679-3@student.aiub.edu}

\begin{document}

\begin{abstract}
Regional linguistic variation remains an important challenge for Bangla natural language processing, particularly in informal and non-standard text. This paper introduces \textbf{BanglaDial-Abuse}, a balanced Bengali-script dataset developed for regional dialect identification in abusive and hostile Bangla text. The dataset contains \textbf{1,000 sentences} distributed equally across four linguistic varieties: \textbf{Standard Bangla, Chattagram, Sylhet, and Barishal}, with 250 samples per class. The resource was constructed using a corpus-grounded synthetic procedure incorporating regional variation in pronouns, possessive forms, verb morphology, negation, interrogative structures, postpositions, vocabulary, and Bengali-script spelling conventions while preserving the underlying hostile or abusive meaning. Descriptive analysis shows broadly comparable sentence-length distributions but partially distinct lexical spaces across the four classes. Pairwise Jaccard vocabulary similarity ranges from 0.37 to 0.56. The primary task is four-class regional dialect identification rather than binary abusive-text detection. The dataset is publicly available through Zenodo under a Creative Commons Attribution 4.0 license. The current version is intended as a research and prototyping corpus rather than a native-speaker-validated gold-standard linguistic resource.
\end{abstract}



\maketitle

\section{Introduction}

Bangla exhibits substantial regional diversity in vocabulary, morphology, pronunciation, grammatical constructions, and informal orthographic representation. Although Standard Bangla dominates many written resources and computational benchmarks, regional forms are frequently encountered in informal communication and online conversational environments. Consequently, models trained primarily on standardized text may encounter difficulty when the same semantic content is expressed using regional pronouns, verbal forms, lexical items, particles, or spelling conventions.

Recent work has begun to provide computational resources for regional Bangla. Mahi et al. introduced \textbf{BanglaDial}, a merged corpus for Bengali regional dialect analysis \cite{mahi2025bangladial}. Its scale makes it useful for general dialect-identification research, but the corpus is not specifically designed around abusive-language contexts. Sultana et al. developed \textbf{ONUBAD}, a parallel regional-language resource supporting conversion from Chittagong, Sylhet, and Barisal forms into Standard Bangla \cite{sultana2025onubad}. The resource focuses primarily on translation between regional and standardized forms. Rahman et al. introduced \textbf{BD-Dialect}, containing aligned Standard Bangla, English, and several regional varieties \cite{rahman2026bddialect}. This work highlights the value of native-speaker collection and validation for regional-language resources. Together, these resources demonstrate increasing interest in dialect-aware Bangla language technologies.

A separate line of research has addressed abusive and hateful Bangla. For example, Romim et al. presented \textbf{BD-SHS}, a benchmark for online Bangla hate speech containing manually labeled offensive Bangla comments collected from different online social contexts \cite{romim2022bdshs}. Such work demonstrates the importance of high-quality abusive-language resources but does not focus on regional dialect recognition. In general, abusive-language resources and regional dialect resources have been developed for different computational objectives. This separation leaves comparatively limited support for investigating how abusive or hostile semantic content is expressed through regional Bangla varieties.

From a modeling perspective, Bangla-specific pretrained language models have also improved support for downstream Bangla NLP. BanglaBERT, for example, was pretrained on a large Bangla corpus and evaluated through the Bangla Language Understanding Benchmark \cite{bhattacharjee2022banglabert}. However, pretrained language models trained predominantly on standardized or broadly collected Bangla may still encounter challenges when processing region-specific lexical, morphological, grammatical, and orthographic patterns, particularly in informal and hostile communication.

These observations motivate the construction of a small controlled resource in which abusive or hostile semantic contexts are represented through multiple regional linguistic varieties. We therefore introduce \textbf{BanglaDial-Abuse}, formally released as the \textit{Bangla Abusive Regional Dialect Dataset} \cite{sifat2026dataset}. The dataset contains 1,000 Bengali-script sentences equally distributed among four linguistic varieties: Standard Bangla, Chattagram, Sylhet, and Barishal. By combining a controlled abusive-language domain with regional linguistic variation, BanglaDial-Abuse complements existing dialect and abusive-language resources and provides a focused benchmark for studying regional dialect identification and dialect-aware Bangla NLP.

The main contributions of this dataset note are:
\begin{itemize}
    \item a balanced 1,000-sample resource covering four Bangla linguistic varieties in an abusive and hostile textual domain;
    \item a corpus-grounded synthetic construction procedure incorporating lexical, morphological, grammatical, and orthographic variation;
    \item quantitative characterization of sentence length, frequent lexical items, and pairwise vocabulary overlap; and
    \item an openly released and DOI-addressable research resource for studying regional dialect identification and dialect-aware Bangla NLP.
\end{itemize}

\section{Dataset Construction and Release}

BanglaDial-Abuse was developed as a \textbf{corpus-grounded synthetic regional-language resource}. The construction objective was to preserve the underlying hostile or abusive meaning of an utterance while expressing it using linguistic forms associated with a target regional variety.

Regional sentence construction was guided by recurring patterns observed in Bangla regional-language material. The process considered variation in:
\begin{itemize}
    \item pronouns and possessive forms;
    \item verb morphology;
    \item negation patterns;
    \item interrogative constructions;
    \item postpositions;
    \item regional vocabulary; and
    \item Bengali-script spelling conventions.
\end{itemize}

The procedure was not intended to operate only through isolated lexical replacement. Multiple regional signals were incorporated where appropriate to represent variation at lexical, morphological, grammatical, and orthographic levels.

The released file contains two fields:
\begin{itemize}
    \item \texttt{text}: an abusive or hostile Bangla sentence;
    \item \texttt{dialect\_label}: the corresponding regional variety.
\end{itemize}
No speaker identity, demographic information, or personally identifiable information is included.

\subsection{Dataset Release}

Version 1.0 was publicly released on October 1, 2026 through Zenodo \cite{sifat2026dataset}. The release contains the file \texttt{bangla\_abusive\_dialect\_text.xlsx} and is distributed under the Creative Commons Attribution 4.0 International license.

The version-specific DOI is:
\begin{center}
\url{https://doi.org/10.5281/zenodo.23074319}
\end{center}

Zenodo also provides the concept DOI \url{https://doi.org/10.5281/zenodo.23074318}, which resolves to the latest available dataset version. Development materials and repository documentation are available at:
\begin{center}
\url{https://github.com/almassifat/Bangla-Regional-Dialect-Abusive-Dataset}
\end{center}

\section{Dataset Characteristics}

The current release contains exactly 1,000 samples. Each of the four classes contributes 250 samples, resulting in a balanced four-class distribution. The released version contains no missing values and no exact duplicate rows.

\begin{table}[t]
\caption{Class distribution of BanglaDial-Abuse Version 1.0.}
\label{tab:distribution}
\centering
\small
\begin{tabular}{lrr}
\toprule
\textbf{Dialect} & \textbf{Samples} & \textbf{Share} \\
\midrule
Standard Bangla & 250 & 25\% \\
Chattagram      & 250 & 25\% \\
Sylhet          & 250 & 25\% \\
Barishal        & 250 & 25\% \\
\midrule
\textbf{Total}  & \textbf{1,000} & \textbf{100\%} \\
\bottomrule
\end{tabular}
\end{table}

\subsection{Sentence-Length Distribution}

\begin{figure}[t]
    \centering
    \includegraphics[width=0.90\columnwidth]{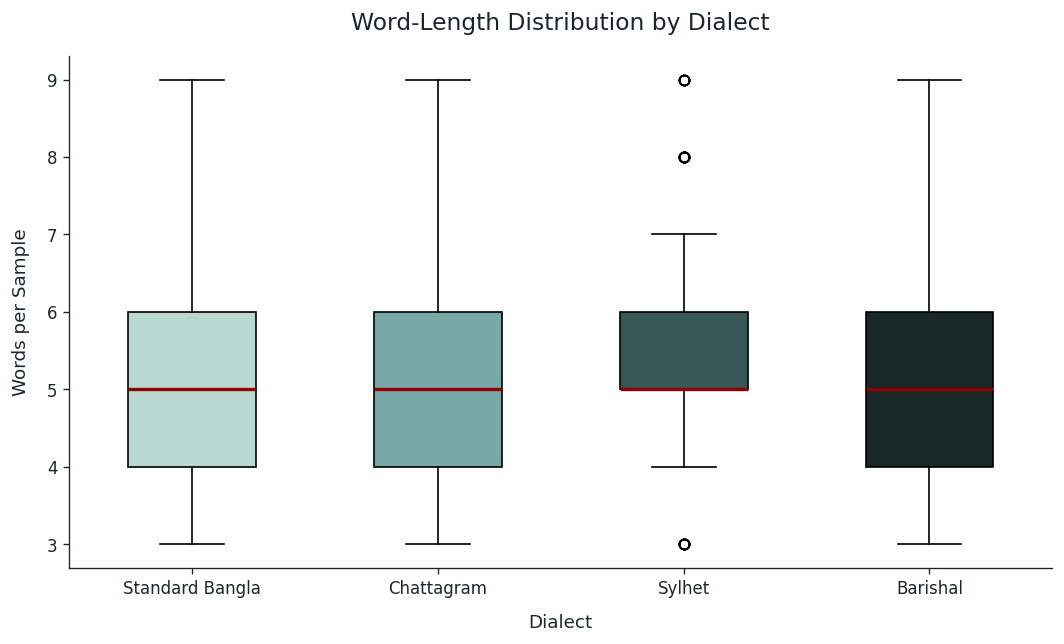}
    \caption{Word-length distribution across the four dialect classes. Median sentence lengths remain broadly comparable across varieties.}
    \label{fig:length}
\end{figure}

The samples primarily consist of short conversational utterances. Figure~\ref{fig:length} summarizes the number of whitespace-separated words per sample.

Across the four classes, the median sentence length is approximately five words. Standard Bangla, Chattagram, and Barishal show broadly similar interquartile ranges, while Sylhet is more concentrated around five to six words, with several shorter and longer observations.

The comparable sentence-length distributions are useful for the intended classification task because sentence length alone is unlikely to serve as a strong trivial separator. Successful classification therefore requires greater reliance on lexical, morphological, or orthographic information.

\subsection{Frequent Lexical Patterns}

\begin{figure*}[t]
    \centering
    \includegraphics[width=0.80\textwidth]{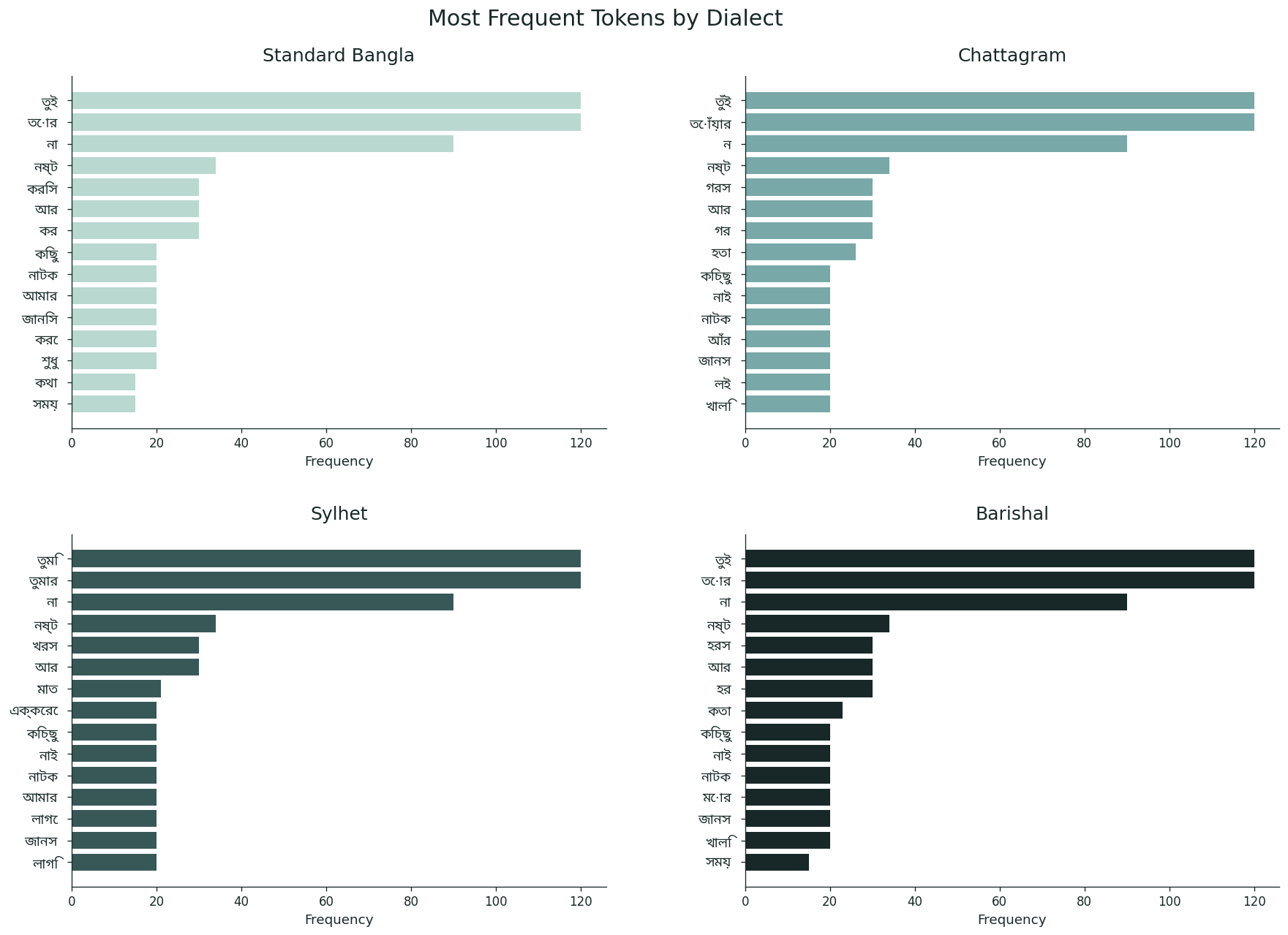}
    \caption{Most frequent tokens in Standard Bangla, Chattagram, Sylhet, and Barishal. Shared high-frequency vocabulary coexists with regional lexical and orthographic variation.}
    \label{fig:tokens}
\end{figure*}

Figure~\ref{fig:tokens} presents the most frequent tokens within each regional class. Several high-frequency items appear across multiple dialects because the dataset represents related conversational and hostile semantic contexts. At the same time, different lexical realizations and orthographic forms occur with different frequencies across the four classes.

This observation suggests that the task is not simply one of recognizing completely disjoint vocabularies. A classifier must identify regional cues within substantial shared lexical material. Character-level representations may therefore be particularly relevant because regional differences may appear as relatively small morphological or spelling changes.

\subsection{Vocabulary Overlap}

Pairwise vocabulary overlap was quantified using Jaccard similarity. Let $V_i$ and $V_j$ denote the sets of unique tokens occurring within two dialect classes. Their similarity is computed as
\begin{equation}
J(V_i,V_j)=\frac{|V_i \cap V_j|}{|V_i \cup V_j|}.
\end{equation}

\begin{figure}[t]
    \centering
    \includegraphics[width=0.82\columnwidth]{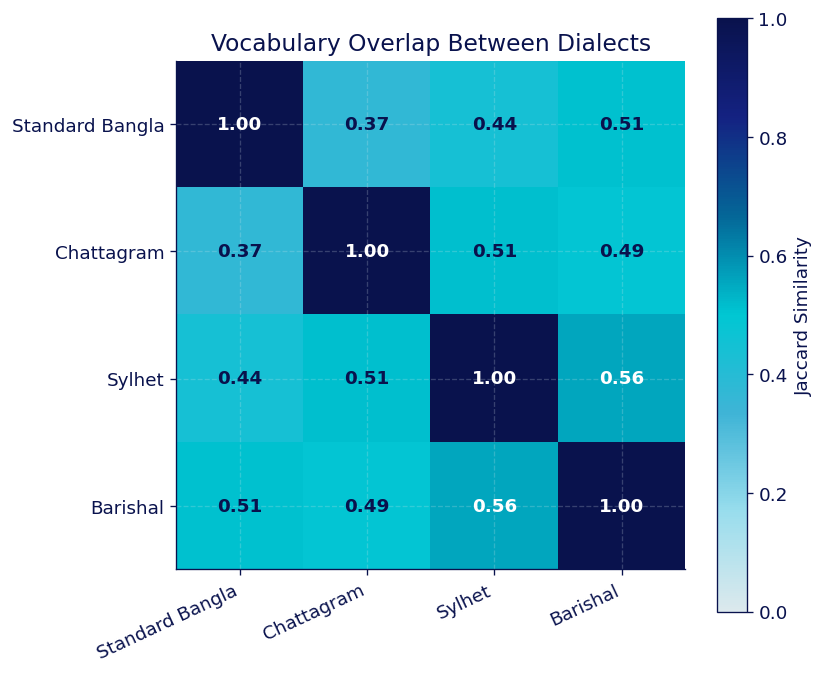}
    \caption{Pairwise vocabulary overlap measured using Jaccard similarity. Sylhet--Barishal has the highest observed overlap (0.56), whereas Standard Bangla--Chattagram has the lowest (0.37).}
    \label{fig:vocab}
\end{figure}

Figure~\ref{fig:vocab} presents the resulting similarity matrix. The highest observed overlap occurs between \textbf{Sylhet and Barishal}, with a Jaccard similarity of \textbf{0.56}. Standard Bangla--Barishal and Chattagram--Sylhet each reach approximately \textbf{0.51}, while Chattagram--Barishal reaches \textbf{0.49}.

Standard Bangla and Sylhet have a similarity of \textbf{0.44}. The lowest observed overlap occurs between \textbf{Standard Bangla and Chattagram}, with a Jaccard similarity of \textbf{0.37}.

These results indicate that the four classes occupy partially overlapping rather than independent lexical spaces. This makes the resource useful for examining whether classification systems identify combinations of regional features rather than relying only on unique words. The values should, however, be interpreted as characteristics of BanglaDial-Abuse itself, not as population-level estimates of lexical similarity among the complete regional varieties.

\section{Task Definition and Intended Use}

The primary task associated with BanglaDial-Abuse is \textbf{multi-class regional dialect identification}. Given an abusive or hostile Bengali-script sentence $x$, the objective is to estimate its regional class:
\begin{equation}
f(x) \rightarrow y,
\end{equation}
where
\begin{equation}
y \in \{\text{Standard Bangla, Chattagram, Sylhet, Barishal}\}.
\end{equation}

The dataset is not intended as a binary abusive-language detection corpus, because all samples already belong to the abusive or hostile textual domain. The prediction target is the regional linguistic variety.

Potential baseline approaches include word-level TF--IDF, character $n$-gram TF--IDF, support vector machines, recurrent or convolutional character models, multilingual Transformer encoders, and Bangla-specific pretrained models such as BanglaBERT \cite{bhattacharjee2022banglabert}.

Suitable evaluation measures include accuracy, macro precision, macro recall, macro F1-score, weighted F1-score, Matthews correlation coefficient, and class-wise confusion matrices.

An important experimental consideration concerns semantically matched regional realizations. When multiple dialect variants originate from the same underlying semantic item, all variants should remain within the \textbf{same training, validation, or test partition}. Grouped splitting reduces the risk that highly related versions of the same utterance appear simultaneously in training and evaluation data.

\section{Potential Research Applications}

BanglaDial-Abuse is intended as a small research resource rather than a large general-purpose corpus. Potential applications include regional Bangla dialect identification, dialect-aware abusive-language processing, cross-dialect robustness evaluation, character-level versus token-level classification, regional lexical and morphological analysis, transfer learning across Bangla varieties, and explainable dialect classification.

An additional research direction is to investigate whether strong classification performance represents meaningful modeling of multiple dialectal signals or merely dependence on a limited number of highly recognizable regional markers. Feature attribution, character $n$-gram analysis, and controlled marker-removal experiments could be used to investigate this issue.

\section{Limitations and Ethical Considerations}

BanglaDial-Abuse Version 1.0 should be considered a \textbf{research and prototyping corpus rather than a native-speaker-validated gold-standard resource}. Its main limitations are the small size of 250 instances per class, the use of corpus-grounded synthetic construction, and the simplification of substantial intra-regional variation within Chattagram, Sylhet, and Barishal. Synthetic generation may introduce unnatural expressions, mixed regional features, orthographic inconsistencies, or overly distinctive dialect markers; therefore, model performance should not be interpreted as direct evidence of robustness on naturally occurring regional language.

The current release is limited to Bengali-script abusive and hostile text and does not include Romanized Bangla, speech, or phonological information. Future versions should incorporate systematic native-speaker validation for naturalness, grammaticality, semantic preservation, abusive-intensity preservation, and orthographic acceptability, together with inter-annotator agreement where applicable.

Because the dataset contains offensive language, it should be used only for legitimate research purposes such as dialect identification, moderation, linguistic analysis, and robustness evaluation. Regional labels represent linguistic varieties and should not be used to infer personal, behavioral, or social characteristics of individual speakers.

\section{Data Availability and Reproducibility}

BanglaDial-Abuse Version 1.0 was published openly on Zenodo on October 1, 2026 \cite{sifat2026dataset}. The release contains 1,000 Bengali-script samples distributed equally across four classes.

\medskip
\noindent\textbf{Version-specific DOI:}\\
\url{https://doi.org/10.5281/zenodo.23074319}

\medskip
\noindent\textbf{All-versions DOI:}\\
\url{https://doi.org/10.5281/zenodo.23074318}

\medskip
\noindent\textbf{Source repository:}\\
\url{https://github.com/almassifat/Bangla-Regional-Dialect-Abusive-Dataset}

\medskip
\noindent\textbf{License:} Creative Commons Attribution 4.0 International (CC BY 4.0).

\section{Conclusion}

This paper introduced \textbf{BanglaDial-Abuse}, a balanced, corpus-grounded synthetic dataset containing 1,000 Bengali-script abusive and hostile sentences representing Standard Bangla, Chattagram, Sylhet, and Barishal. Each class contains 250 samples, providing a controlled setting for preliminary regional dialect-classification experiments.

Descriptive analysis shows broadly comparable sentence lengths across the four classes while revealing both shared and regionally distinctive lexical patterns. Pairwise Jaccard vocabulary similarity ranges from 0.37 between Standard Bangla and Chattagram to 0.56 between Sylhet and Barishal, illustrating that the classes contain substantial shared vocabulary while retaining distinguishable surface characteristics.

\end{document}